\documentclass{ifacconf}

\usepackage[round]{natbib} 
\usepackage{graphicx}      
\usepackage{subfigure}
\usepackage{enumerate}
\usepackage{amsmath}
\usepackage{amssymb}
\usepackage{booktabs}
\usepackage[english]{babel}
\usepackage{color}

\usepackage{moreverb}
\usepackage[english]{babel}
\usepackage{widetext}
\usepackage[]{graphicx}
\usepackage{amsmath}
\usepackage{color}
\usepackage{mathptmx} 
\usepackage{amssymb}  
\usepackage{epstopdf}
\usepackage{tikz}

\usepackage{graphicx}      
\usepackage{natbib}        

\graphicspath{{Figures/}}

\begin{document}
\begin{frontmatter}

\title{Tustin neural networks: a class of recurrent nets for
 adaptive MPC of mechanical systems}


\author[First,Second]{S. Pozzoli}
\author[First]{M. Gallieri}
\author[Second]{R. Scattolini}

\address[First]{NNAISENSE,  Lugano, Switzerland\\(e-mail: name@nnaisense.com)}
\address[Second]{Dipartimento di Elettronica, Informazione e Bioingegneria, Politecnico di Milano, via Ponzio 34/5, 20133 Milano, Italy\\ (e-mail: name.surname@polimi.it)}


\begin{abstract}:                 
 The use of recurrent neural networks to represent the dynamics of unstable systems is difficult due to the need to properly initialize their internal states, which in most of the cases do not have any physical meaning, consequent to the non-smoothness of the optimization problem. For this reason, in this paper focus is placed on mechanical systems characterized by a number of degrees of freedom, each one represented by two states, namely position and velocity. For these systems, a new recurrent neural network is proposed: Tustin-Net. Inspired by second-order dynamics, the network hidden states can be straightforwardly estimated, as their differential relationships with the measured states are hard-coded in the forward pass.  The proposed structure is used to model a double inverted pendulum and for model-based Reinforcement Learning, where an adaptive Model Predictive Controller scheme using the Unscented Kalman Filter is proposed to deal with parameter changes in the system.
 \end{abstract}


\end{frontmatter}
%
\section{Introduction}\label{Intro}
Recurrent Neural Networks (RNNs), like \textit{Echo State Networks} (ESNs) (\cite{jaeger2002tutorial}, \cite{armenio2019model}), and \textit{Long-Short Term Memory} networks  (LSTMs) (\cite{hochreiter1997long, gers1999learning, greff2017lstm}), are nowadays widely used in many engineering fields, such as speech recognition (\cite{tai2015improved,greff2017lstm}), mobile phones, and GPS navigators (\cite{zhao2019traffic}). In addition, they have the potentiality to be successfully applied for control of dynamic systems, since they allow to obtain reliable dynamical models of the system directly from data, so reducing the time and effort required to develop physical-based models. In turn, RNNs models can then be used in model-based control design techniques, like Linear Quadratic or Model Predictive Control (MPC, \cite{Macie:2002, rawlings2009model}). However, the potentialities of RNNs as estimated dynamic models for control design has not been fully exploited yet, and some fundamental properties, like stability, have not been studied in depth. Among the main contributions concerning ESNs, their incremental Input-to-State Stability ($\delta ISS$) property (\cite{Discrete-timedeltaISS}) has been analyzed in \cite{armenio2019model}, where also the design of stable observers and stabilizing MPC laws have been considered. Local stability analysis and equilibria computation of LSTMs have been addressed in (\cite{stipanovic2018some}, \cite{amrouche2018long}), while their $\delta ISS$ has been studied in (\cite{terzi2019model}). in addition, the industrial application of LSTM for control has been discussed in (\cite{lanzetti2019recurrent}).

The use of RNNs in the modeling of dynamical \emph{unstable} systems still remains a largely unexplored field, due to the fact that the most popular and previously mentioned structures can still fail to provide satisfactory results. This is due to the fact that, for unstable systems, gradient descent can become ill-posed (\cite{doya_bifurcations_1993, pascanu_understanding_2012, Ciccone2018NAISNetSD}) due to the vanishing and exploding gradient problem (\cite{hochreiter1998recurrent}). The possibility for bifurcation and exploding gradients implies, for instance, that the proper initialization of the network internal states is crucial to achieve satisfactory results. Unfortunately, RNNs are typically characterized by a large number of hidden states, not related to any physical properties of the system, so that their initialization cannot rely on any a-priori physical knowledge. The networks are usually trained via stochastic gradient descent on batches of data, for instance using methods like \emph{Adam} (\cite{kingma_adam:_2014}). These methods require a large amount of data as well as parameters in order to effectively exploit the benefits of stochastic search. Moreover, they are not suited for online refinement of the model as new data comes in (\cite{ash_difficulty_2019}). This limits the applicability of neural network models to scenarios that are either stationary or where dynamics are slow/stable enough so that a new model can be trained from scratch without affecting the system safety.

Adaptive control (\cite{Astrom:1994:AC:546778}) is a branch of control theory that deals with changing environments with the aim of safely refining the control policy or the model in order to improve performance. In the context of MPC, \cite{marafioti_persistently_2014, heirung_mpc-based_2015} incorporated conditions on the input signals such that the dynamics can be best adapted. \cite{lorenzen_cannon} formulated a stable adaptive MPC approach based on an  tube approach which restricts constraints adaptively while refining a model uncertainty set estimate. \cite{Adaptive_MPC_EKF} adapted NN models using the Extended Kalman Filter (EKF \cite{anderson2012optimal}).

In this paper, we propose a new network structure suitably developed to represent unstable dynamical systems, with particular focus on robotic systems. The main idea behind the development of this network structure is to fully exploit the knowledge of the relationships between variables, typically positions and velocities (linear or angular), and to represent their dependence on exogenous variables, like forces and torques, by means of suitable feedforward networks. By using this network, the initialization problem can be trivially solved, and the gradient flow is facilitated so that training is less affected by
vanishing (and exploding) gradients. Being based on the Tustin and Euler discretization methods (\cite{aastrom2013computer}), they will be named \emph{Tustin-Nets} (TN). Secondly, we propose an adaptive MPC scheme that makes use of the Unscented Kalman Filter (UKF, \cite{wan2000unscented}) to efficiently control a changing environment. In particular, the last layer of the network is updated online using the UKF while simultaneously estimating the network hidden states. The approach is tested on a double pendulum with changing dynamical properties.

The characteristics of TNs are described in the following Section \ref{SectTEN}, while their use for control of a double inverted pendulum is considered in Section \ref{eq: TENtrain}. An adaptive implementation of the proposed control scheme is reported in Section \ref{aMPC}, while some conclusions and hints for future work are listed in Section \ref{Conc}.
\section{Tustin-Nets}\label{SectTEN}
Mechanical systems have a certain number of degrees of freedom (DOF), for example joints' angles or translations, each one represented by two states: its position $\theta$ and its velocity $\dot{\theta}$. Due to the presence of inertia, the velocity does not vary instantaneously and depends on the applied torque, on the position, and on the velocity itself. The relation between velocities and positions is known a priori and can be approximated by means of the \textit{Tustin} and/or \emph{Euler} approximation. Specifically, denoting by $s$ and $z$ the complex variables referred to the \emph{S-} and \emph{Z-} planes for continuous time and discrete time systems, respectively, the Tustin (TU) and (forward) Euler (EU) transformations read:
$$s=\frac{2}{T_s} \frac{z+1}{z-1}\ (TU) \qquad , \qquad s=\frac{z-1}{T_s}\  (EU)$$
where $T_s$ is the adopted sampling time.\\
Now, assume that the mechanical system referred to a generic DOF is described by the following second-order model:
\begin{eqnarray} \label{eq: tustin net general equations1}
    \ddot{x}(t) &=& f(x(t),\dot{x}(t),u(t)) \\ \nonumber
    y(t) &=& x(t)
\end{eqnarray}
where $t$ is the continuous time index, $x$ and $\dot{x}$ are the position and the velocity, respectively, and $u$ is the (vector) torque applied. In (\ref{eq: tustin net general equations1}) we assume that measurements, $y(t)$ include only the position. By applying the TU transformation to the first state equation and the EU transformation to the second, one obtains at sampling times $k$
\begin{eqnarray} \label{eq: TENform}
    {x}(k+1) &=& {x}(k) + T_s \left(\frac{\dot{{x}}(k+1)+\dot{{x}}(k)}{2}\right)\\ \label{eq: tustin net general equations}
    \dot{{x}}(k+1) &=& \dot{x}(k) + T_s f\left({x}(k),\dot{x}(k),u(k)\right)\\ \nonumber
    {y}(k) &=& {x}(k)
\end{eqnarray}
System \eqref{eq: TENform} represents the basic brick of TNs, where the function $f$ can be described by a feedforward NN to be trained with experimental data and $x$, $\dot{x}$ are vectors of suitable dimensions.
According to their definition, initialization of the variables $x_{i}$, $\dot{x}_{i}$ can be trivially performed, for instance using the Euler differentiation.
\section{Inverted pendulum: modeling, estimation, and control}\label{eq: TENtrain}
\begin{figure}
    \centerline{\includegraphics[width=0.3\linewidth]{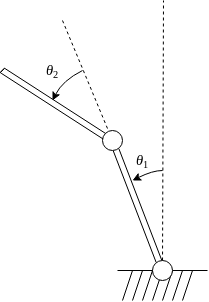}}
    \caption{{{\bf Double inverted pendulum schematics.} Both joints are torque actuated with limited ranges. Velocities are assumed not measurable and are estimated.}}    \label{fig:double_pendulum}
\end{figure}

The double inverted pendulum, represented in Figure \ref{fig:double_pendulum}, is a popular benchmark to test new control techniques due to its high nonlinearity and instability. A double pendulum is a mechanical system consisting in two beams connected by a hinge. The beams are free to rotate and their positions are controlled applying a torque to both joints. The system has multiple equilibrium points: three unstable equilibria and a stable one. The goal is to control the double pendulum in its unstable position where both the state variables, i.e. the angles $\theta_1$ and $\theta_2$ are null.

\subsection{Lagrangian modeling of the double pendulum}
The Lagrangian modeling approach has been used to get a physics-based model in which the user can set the friction coefficients of the joints, can modify the mass and the length of the two beams and set the position of their center of mass along their main direction. This model has been used as simulator to test the effectiveness of the developed estimation and control algorithms.
 By means of the Lagrange equation, \cite{pozzoli} formulated the pendulum dynamics as:
\begin{equation*}
    \left[\begin{array}{c}
        \ddot{\theta}_1 \\
        \ddot{\theta}_2
    \end{array}\right] = A(\theta, \dot{\theta})^{-1}B(\theta, \dot{\theta}, u)
\end{equation*}
where the matrices $A$ and $B$ are specified in \cite{pozzoli}, together with the system's parameters.
The ODE is solved using a $5$-th order Runge-Kutta method   \citep{dormand_family_1980}.
\subsection{Tustin-Net model}
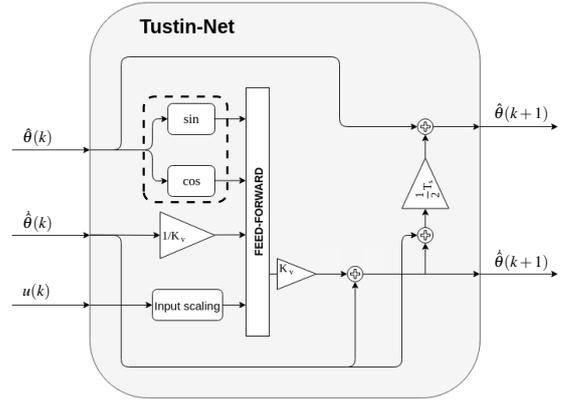
\begin{figure}
\centering
\begin{tikzpicture}
\node[anchor=south west,inner sep=0] at (0,0) { \includegraphics[width=0.9\linewidth]{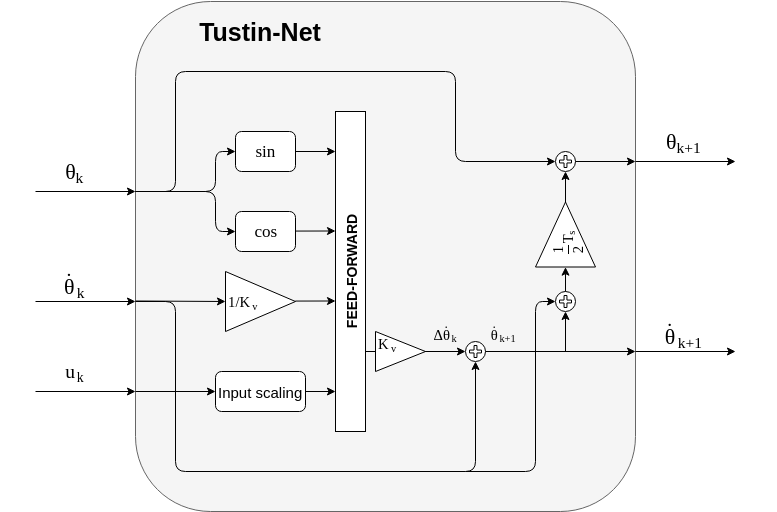}};
\filldraw[white] (0.6,2.) rectangle (1.,1.3) node[anchor=west]{};
\filldraw[white] (0.6,2.2) rectangle (1., 2.6) node[anchor=west]{};
\filldraw[white] (0.6,3.35) rectangle (1., 3.6) node[anchor=west]{};
\filldraw[white] (6.8,3.65) rectangle (7.3, 4.) node[anchor=west]{};
\draw[black, dashed, thick, rounded corners] (2.1,2.6) rectangle (3.2, 4.) node[anchor=west]{};
\filldraw[white] (6.8,1.68) rectangle (7.3, 2.) node[anchor=west]{};
\filldraw[gray!7] (4.4,1.7) rectangle (4.7, 2.) node[anchor=west]{};
\filldraw[gray!7] (5,1.7) rectangle (5.3, 2.) node[anchor=west]{};
\node[anchor=south west,inner sep=0] at (0.5,1.3) {\tiny $u(k)$};
\node[anchor=south west,inner sep=0] at (0.5,2.2) {\tiny$\dot{\hat{\theta}}(k)$};
\node[anchor=south west,inner sep=0] at (0.5,3.33) {\tiny$\hat{\theta}(k)$};
\node[anchor=south west,inner sep=0] at (6.7,3.65) {\tiny$\hat{\theta}(k+1)$};
\node[anchor=south west,inner sep=0] at (6.7,1.68) {\tiny$\dot{\hat{\theta}}(k+1)$};

\end{tikzpicture}
    \caption{{\bf Tustin-net schematics for next-step prediction.} Skip connections facilitate gradient flow. In particular, the input $u$ at time $k$ jointly acts on $\theta$ and $\dot{\theta}$.  Depicted state transformations (dashed box) are a design choice specific to robotic scenarios. }
    \label{fig:tustin_net}
\end{figure}

Different network structures have been tested on the double pendulum, among them LSTMs and ESNs, which however failed to produce satisfactory results due to the difficulty related to their state initialization, which is a crucial step to correctly identify the system dynamics. Therefore, attention has been focused on TNs as defined in \eqref{eq: TENform}, which, in the considered problem can be given the form
\begin{eqnarray}\label{eq: tustin net eq double pendulum} \nonumber
    \hat{\theta}_1(k+1) &=& \hat{{\theta}}_1(k) + T_s K_v\left(\frac{\dot{\hat{\theta}}_1(k+1) +  \dot{\hat{\theta}}_1(k)}{2}\right) \\ \nonumber
    \dot{\hat{\theta}}_1(k+1) &=& \dot{\hat{\theta}}_1(k) + f_1\left(\hat{\theta}_1(k), \dot{\hat{\theta}}_1(k), \hat{\theta}_2(k), \dot{\hat{\theta}}_2(k), u(k)\right) \\
    \hat{\theta}_2(k+1) &=& \hat{\theta}_2(k) + T_s K_v\left(\frac{\dot{\hat{\theta}}_2(k+1) + \dot{\hat{\theta}}_2(k)}{2}\right) \\  \nonumber
    \dot{\hat{\theta}}_2(k+1) &=& \dot{\hat{\theta}}_2(k) + f_2\left(\hat{\theta}_1(k), \dot{\hat{\theta}}_1(k), \hat{\theta}_2(k), \dot{\hat{\theta}}_2(k), u(k)\right) \\ \nonumber
    \hat{y}(k) &=& \hat{\theta}(k)=\left[\begin{array}{c}
        \hat{\theta}_1(k) \\
        \hat{\theta}_2(k)
    \end{array} \right]
\end{eqnarray}
where $\hat{\theta}_i$ and $\dot{\hat{\theta}}_i$ denote the estimates of angles and velocities, and $f_1(\cdot)$ and $f_2(\cdot)$ are nonlinear functions that must be learned and represent how much the velocities have changed with respect to the previous time-step. These are modelled as a 2-hidden-layer fully connected network, with \emph{tanh} activation function and $100$ units, plus a final linear layer. The output of the TN are the measured variables, that coincide with the angular positions. A scheme of this network used for the identification  is depicted in Figure \ref{fig:tustin_net}. For this application, the angles have being used in the form of $\sin$ and $\cos$ features. The coefficient $K_v$ is the ratio between position and velocity normalisation constants. The state and input training data is normalised to have unitary magnitude. In particular, angles and velocities are scaled, respectively, by $\pi$ and $2\pi$, hence $K_v=2$.

 The training and validation dataset include different operating conditions. Specifically, we first consider open loop simulations, where the pendulum starts from different initial positions and then falls under the action of  random input torques. The aim of this is to collect data describing the behavior of the system near the stable equilibrium. Then, closed loop simulations are considerd where a full-state information LQR is used. A stabilising controller is in fact required in order to collect long enough sequences and to correctly capture the dynamics near the unstable equilibrium in the upright position. Furthermore, in order to properly excite the system, during closed loop simulations a white noise is added to the LQR torques. All experiments have a total duration of $12s$, the first $6s$ used for the training and the last $6s$ used for validation. The sampling time is always $T_s = 0.01s$. The initial state of each sequence is retrieved from the available data: the angles are directly the first measurement, while the initial velocities is obtained with a forward Euler approximation. 
%
%
In the training procedure, the cost function to be minimized has been modified from the usual Mean Square Error between the predicted angles and the measured ones to the mean of the square of the difference of \textit{sine} and \textit{cosine} of the angles:
\begin{equation} \label{cost_function}
    {L} = \frac{1}{N}\displaystyle\sum_{k=1}^{N}{\left[(\sin\hat{\theta}(k) - \sin\theta(k))^2 + (\cos\hat{\theta}(k) - \cos\theta(k))^2\right]}
\end{equation}
which, with simple manipulations, can be rewritten as:
\begin{equation} \label{modified_cost_function}
    {L} = \frac{1}{N}\displaystyle\sum_{k=1}^{N}{2\left[1 - \cos(\hat{\theta}(k) - \theta(k)) \right]}
\end{equation}
The network is implemented in Tensorflow \citep{tensorflow2015-whitepaper} and trained using the Adam optimizer \citep{kingma_adam:_2014}.

\begin{figure}
\centering
       \includegraphics[scale=0.33]{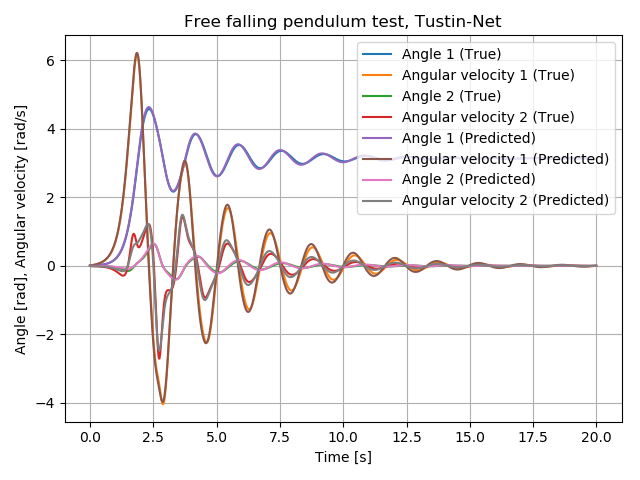}
    \caption{{\bf Open loop predictions.} Double pendulum free falling test.  Predictions from Tustin-Net are long-term accurate when converging to the stable equilibrium.}
   \label{fig:free_falling_test_pendulum_tustin}
\end{figure}

Figure \ref{fig:free_falling_test_pendulum_tustin} shows a comparison of the measured angles' positions and velocities and the corresponding outputs of the trained net in the free falling test. These results highlight the ability of the network to recognize the system equilibria. In fact, as it can be seen, the predicted trajectory follows closely the measured one and asymptotically tends to the stable configuration in $\theta_1=\pi$ and $\theta_2=0$.

\begin{figure}
\centering
       \includegraphics[scale=0.33]{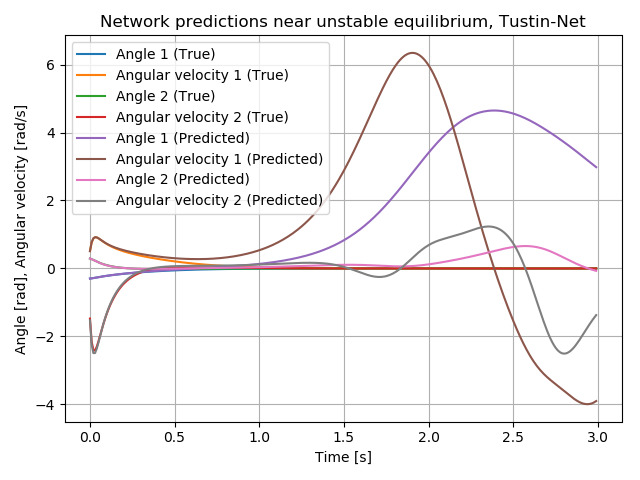}
    \caption{{\bf Closed loop predictions}. Tustin-net predicts the double pendulum states near the unstable equilibrium with reasonable accuracy for the first $0.5s$ (50 steps) when the system is controlled with a full-state information LQR.}
   \label{fig:pendulum_preds_unstable_tustin}
\end{figure}

The results of the closed loop test are reported in Figure \ref{fig:pendulum_preds_unstable_tustin}. Due to the unstable dynamics of the system around the origin, even a small estimation error causes a continuously growing behavior of the variables, so that the predictions provided by the network diverge from the corresponding measurements. However, in the initial period of the simulation, i.e. up to time \emph{0.5s}, the network is able to provide predictions with satisfactory accuracy. Since in the control design phase a much shorter prediction horizon has been used, the accuracy of the model has been judged to be totally adequate for control design.\\
\subsection{TN state estimation and model predictive control}
\begin{figure}
    \centering
    \includegraphics[width=\linewidth]{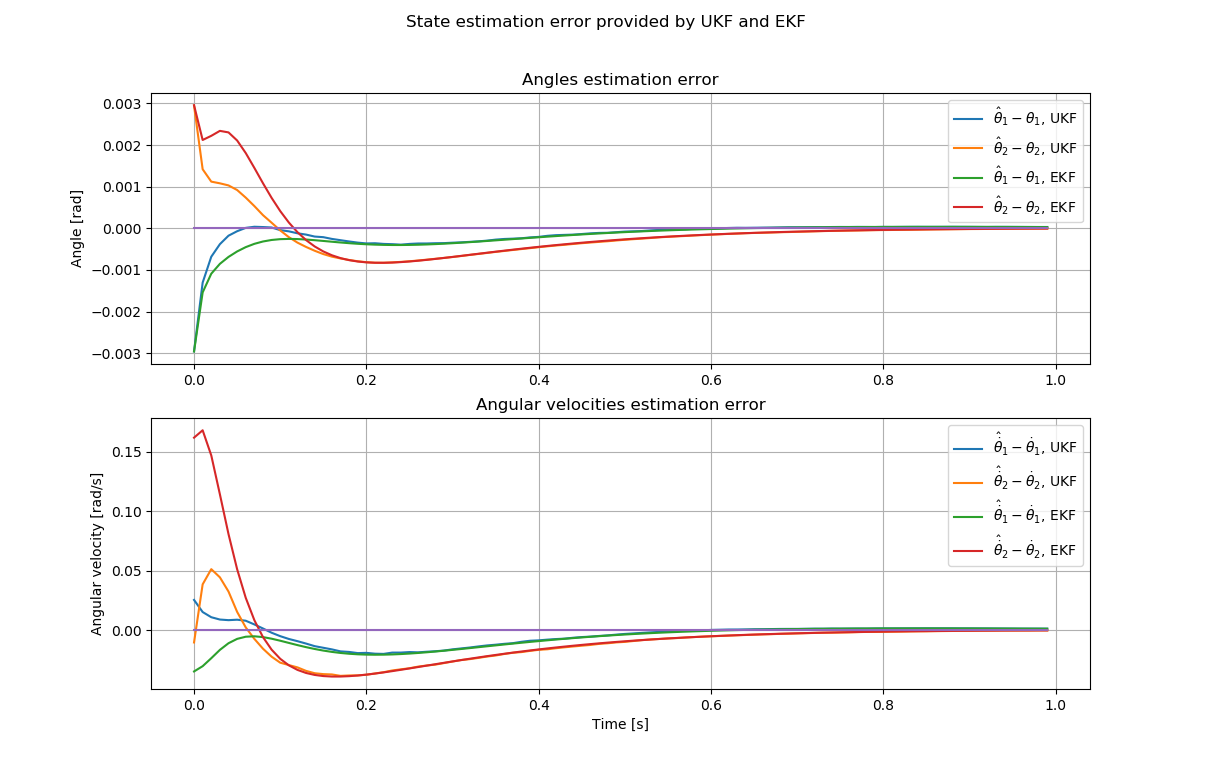}
    \includegraphics[width=\linewidth]{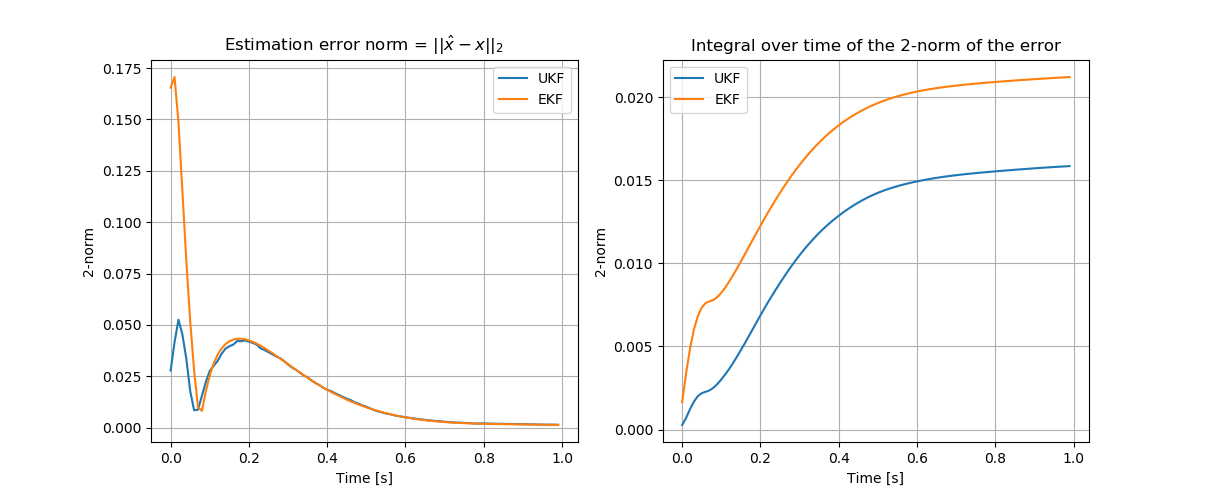}
    \caption{{\bf Comparison of the UKF and the EKF}. With appropriate care for the sigma points location, the UKF significantly outperforms the EKF in terms of estimation error.}
    \label{fig:UKF vs EKF convergence and 2-norm}
\end{figure}

Since our goal is to control the pendulum in its unstable position and the only available measurements are the joints' positions, a state estimator is required to avoid the divergence of the predicted states. Two state estimation algorithms are considered: the Extended Kalman Filter (EKF,  \cite{anderson2012optimal}) and the Unscented Kalman Filter (UKF, \cite{wan2000unscented}). A detailed analysis of the implementation and tuning of these filters is reported in \cite{pozzoli}. Here a comparison of the state estimation error with the two algorithms is illustrated in Figure \ref{fig:UKF vs EKF convergence and 2-norm}, which shows that the state estimate integral RMSE significantly improves using the UKF. A disadvantage of the UKF is that careful selection of the sigma points location is needed for it to be successful (see \cite{Turner2010}). This is further analysed in \cite{pozzoli}. Since the UKF is implemented in Tensorflow (\cite{tensorflow2015-whitepaper}), it is in possible to \emph{learn} the sigma points with gradient-based methods. This is left for future work.

\begin{figure}
    \centering
        \includegraphics[scale=0.3, clip, trim={1.0cm, 0cm, 0cm, 0cm}]{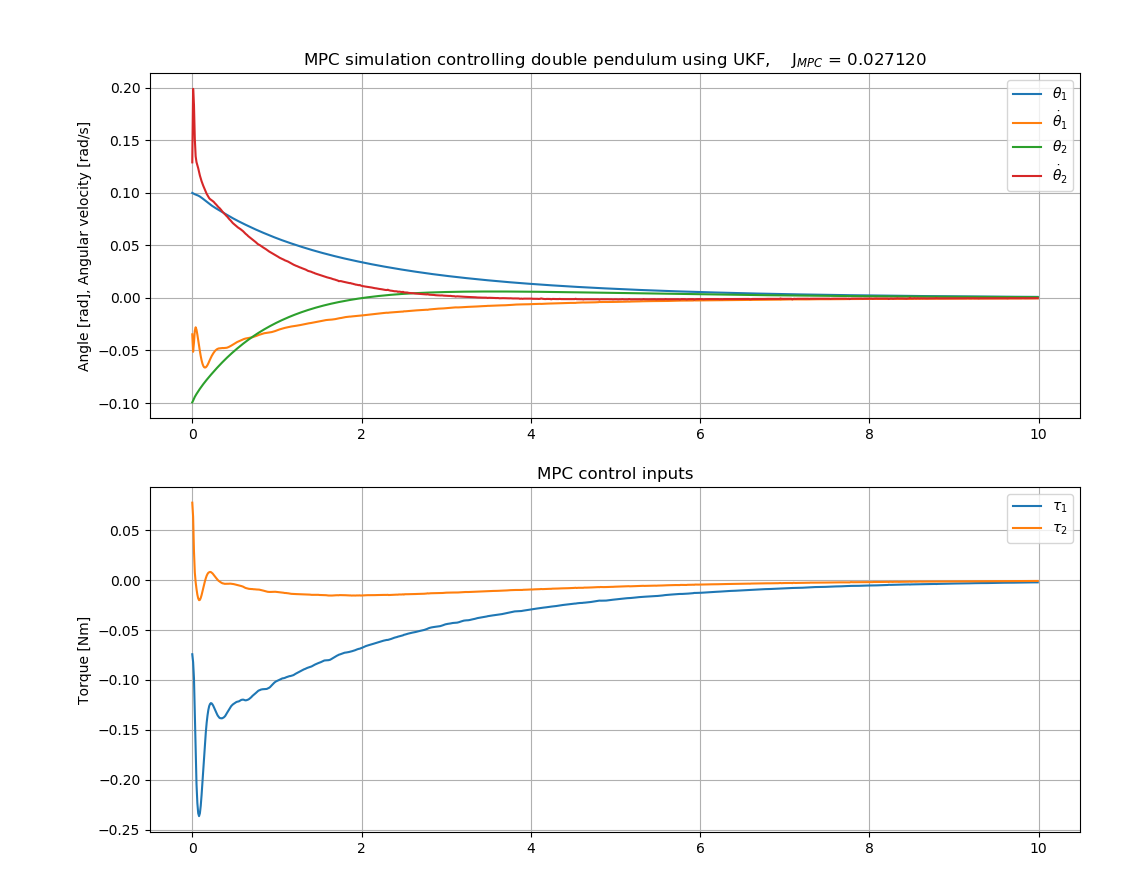}
    \caption{{\bf Tustin-Net MPC results.} For fixed system parameters the learned TN-based UKF and MPC successfully swing-up the double pendulum. }
    \label{fig:MPC_UKF}
\end{figure}

The pendulum control is performed using \emph{Model Predictive Control} (MPC, \cite{Macie:2002,rawlings2017model}). In particular, given the state estimates from the UKF or EKF, the forward model prediction is used to compute a sequence of optimal actions. Then, the first one will be applied and the whole procedure repeated when new measurements come in. Specifically, a prediction horizon of 5 steps, and total length of $0.05s$, has been adopted. The control variables, i.e. the torques, have been saturated, and in the cost function to be minimized the square of the predicted state errors and of the future control moves have been weighted. The control action, computed with the MPC regulator based on the TN model, has been applied to the physical-based simulator of the system. The transients of the state variables, starting from the initial values $[0.1,0,-0.1,0]$, are  reported in Figure \ref{fig:MPC_UKF}.
\section{Neural adaptive MPC}\label{aMPC}
\begin{figure}[h]
    \centering
    \includegraphics[width=0.7\linewidth]{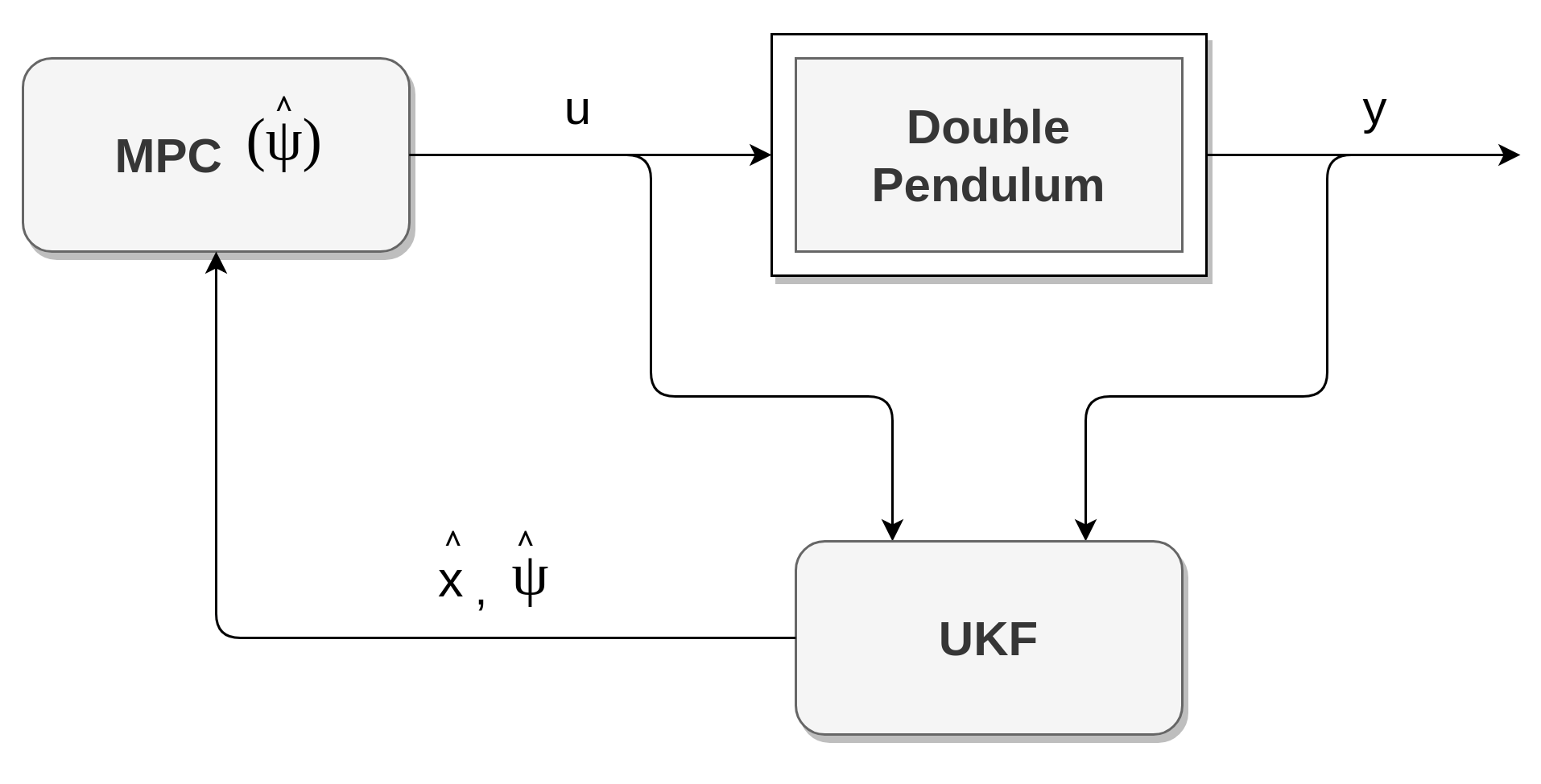}
    \caption{{\bf Closed loop adaptive MPC scheme.} The UKF learns the TN  last layer, additionally to estimating velocities, in order to capture the variations in the system properties.}
    \label{fig:closed loop adaptive MPC scheme}
\end{figure}

The performance of the control approach previously described can seriously deteriorate due to modeling errors caused, for example, by a limited amount and quality of  data available to train the neural network in all its working conditions, or to variations of the physical parameters of the system. An obvious idea to improve the accuracy of the model is to adjust its parameters in an online fashion by formulating a so-called dual identification/control problem. We propose the use of a \textit{joint UKF} for the estimation of the network state and parameters. In particular, the functions $f_1(\cdot)$ and $f_2(\cdot)$ from (\ref{eq: tustin net eq double pendulum}) are adapted by updating their last linear layer. Assuming that the parameters, $\psi$, are constant (or slowly varying), the enlarged model to be estimated is described:
\begin{eqnarray} \label{estate} \nonumber
 \hat{x}(k+1) &=& \hat{x}(k) + T_s \left(\frac{\dot{\hat{x}}(k+1)+\dot{\hat{x}}(k)}{2}\right) +  v_x(k)\\
    \dot{\hat{x}}(k+1) &=& \dot{\hat{x}}(k) + T_s f\left(\hat{x}(k),\dot{\hat{x}}(k),u(k), \psi(k)\right) + v_{\dot{x}}(k)\\ \nonumber
    \psi(k+1) &=& \psi(k) + v_\psi(k) \\ \nonumber
    \hat{y}(k) &=& g(\hat{x}(k)) + n(k)
\end{eqnarray}
where $g$ is the estimated map between states and outputs (in the pendulum case this is just a selector for the positions). Note that, in (\ref{estate}), the vectors $v_x(k)$, $v_{\dot{x}}(k)$, $v_\psi(k)$, and $n(k)$ are assumed to be zero-mean white noises with known variance.
The state estimate, $\hat{x}$, and the parameters $\hat{\psi}$ are recursively updated by the \emph{jUKF} and sent to the MPC algorithm computing the control action, as depicted in Figure \ref{fig:closed loop adaptive MPC scheme}.

\begin{figure}[h]
    \centering
     \includegraphics[scale=0.3, clip, trim={1.0cm, 0cm, 0cm, 0cm}]{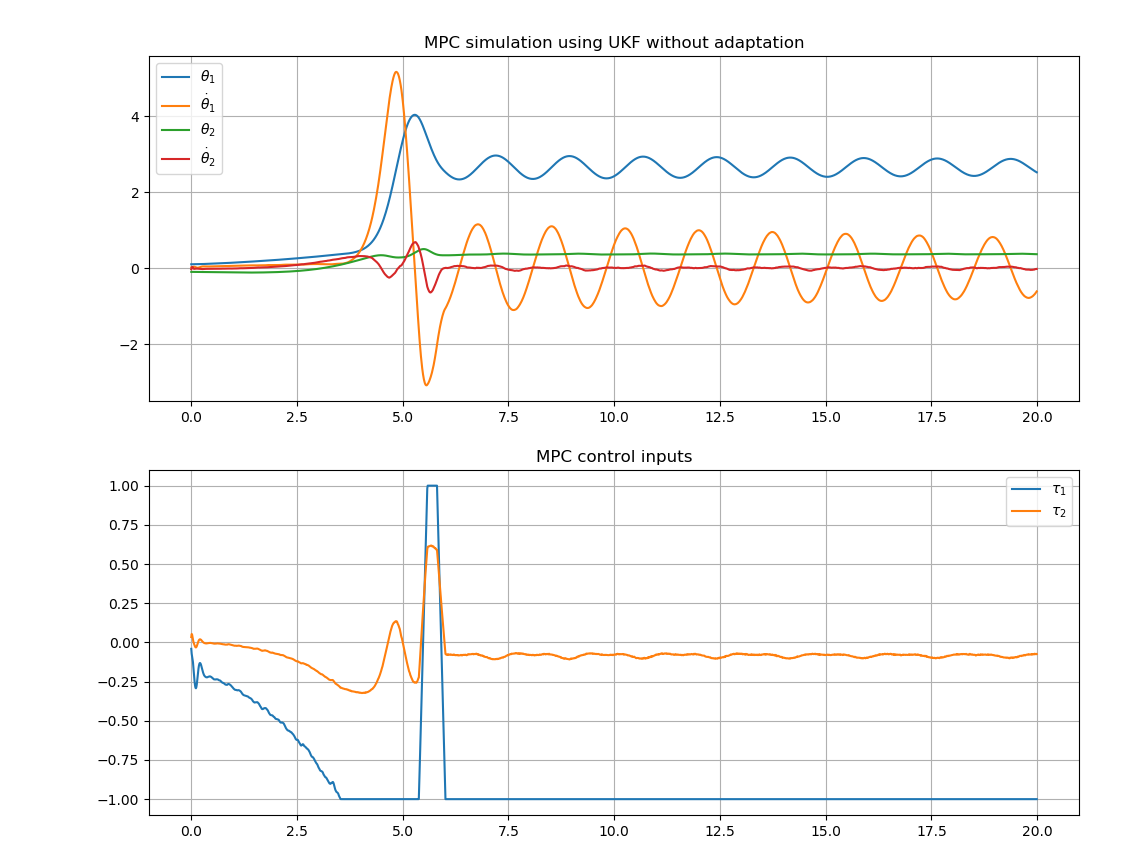}
    \caption{{\bf Different system with non-adaptive MPC.} If the pendulum's physical parameters change then the control performance significantly degrades without adaptation.}
    \label{fig: non adaptive MPC}
\end{figure}

\begin{figure}
    \centering
     \vspace{-1cm}
     \includegraphics[scale=0.3, clip, trim={1.0cm, 0cm, 0cm, 0cm}]{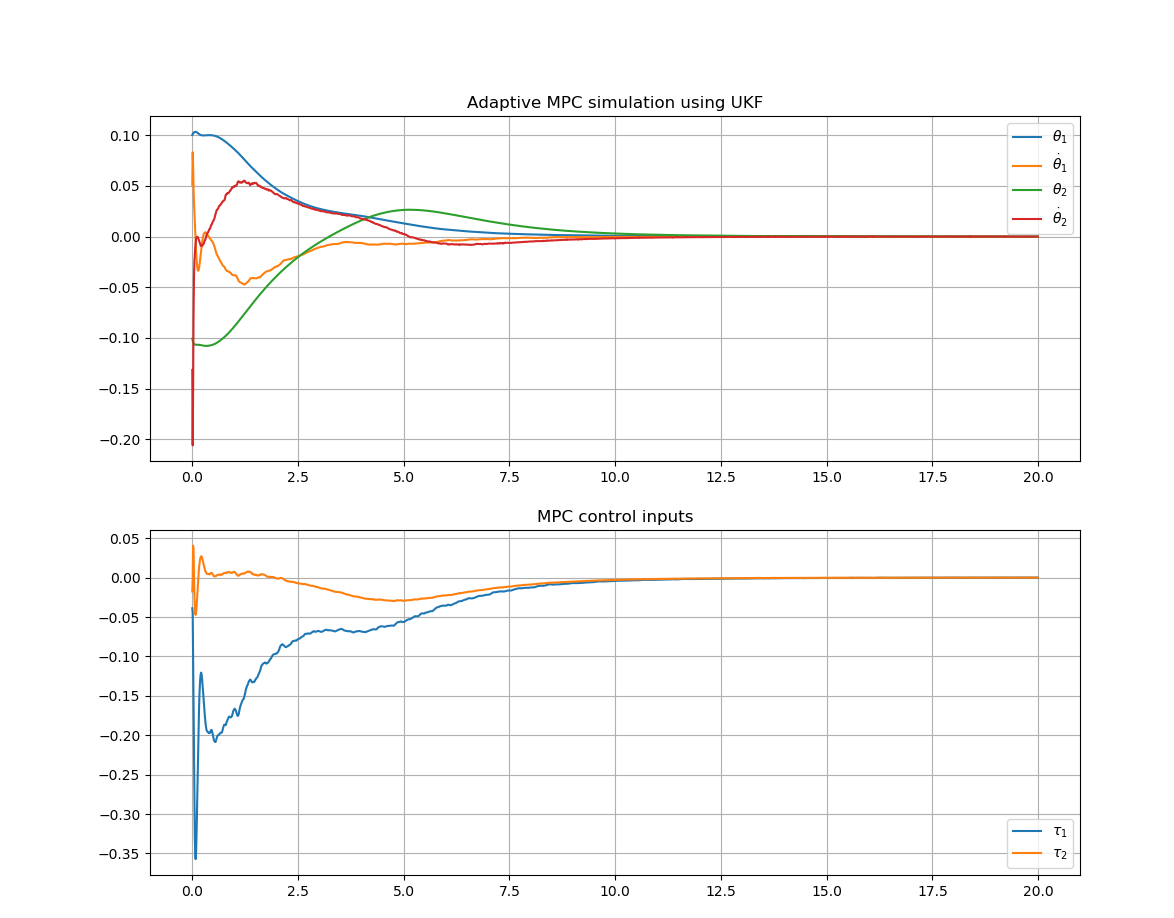}
    \caption{{\bf Neural adaptive MPC.} The UKF learns the TN last layer to capture the different system properties. Performance is recovered.}
    \label{fig:Adaptive MPC}
\end{figure}

\begin{figure}
    \centering

     \includegraphics[scale=0.3, clip, trim={1.0cm, 0cm, 0cm, 0cm}]{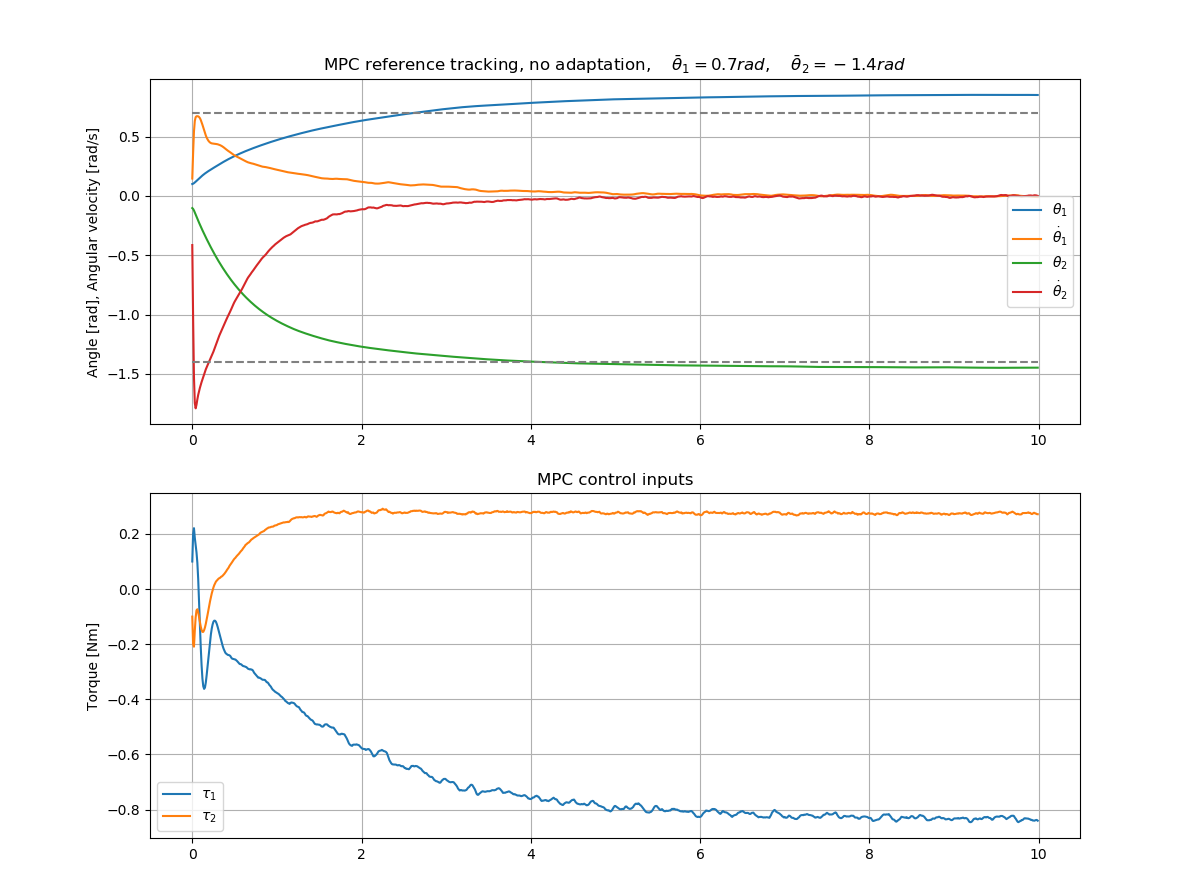}
    \caption{{\bf Reference tracking without adaptation.} The approach is not capable of tracking a non-equilibrium target without adaptation due to model mismatch.}
    \label{fig: MPC_ref_tracking_pendulum}
\end{figure}

\begin{figure}
    \centering
     \vspace{-0.5cm}
    \includegraphics[scale=0.3, clip, trim={1.0cm, 0cm, 0cm, 0cm}]{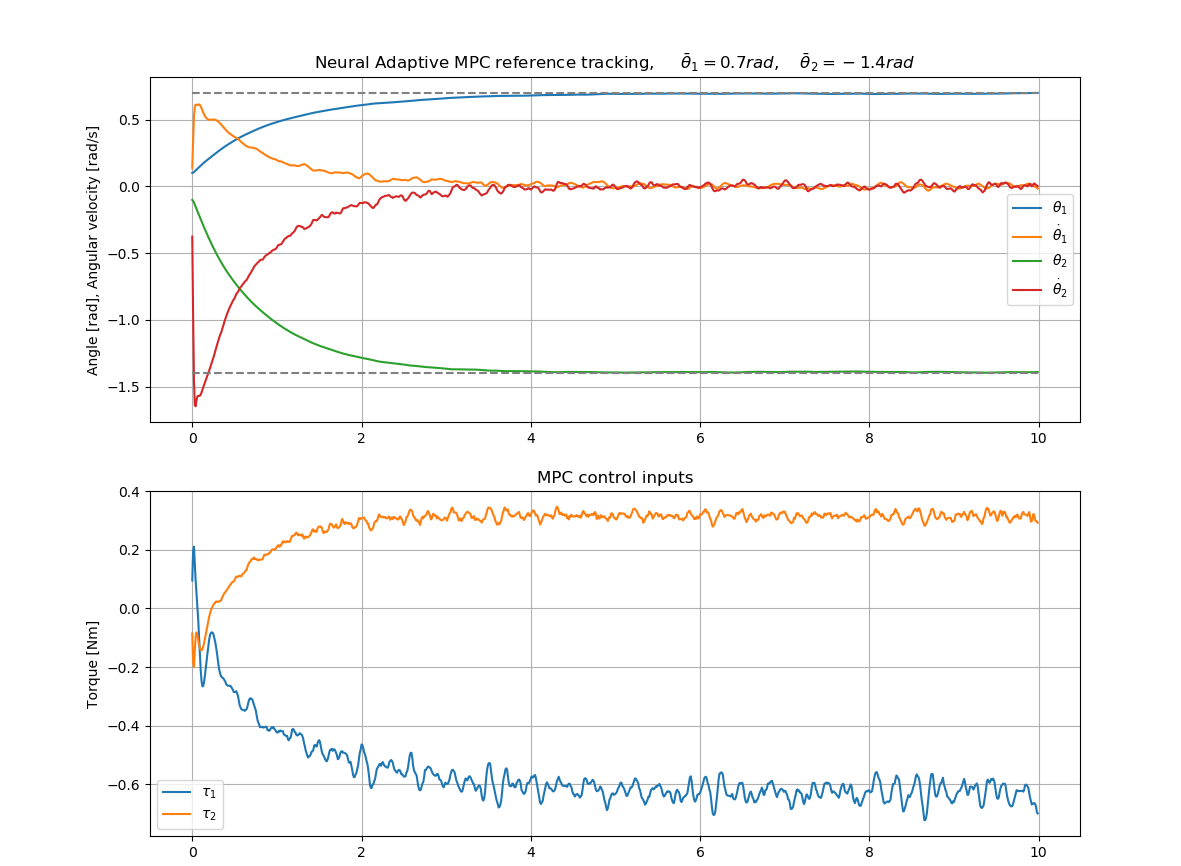}
    \caption{{\bf Adaptive reference tracking TN MPC.} The adaptive TN MPC scheme can track reachable non-equilibrium targets with minimum offset.}
    \label{fig: NAMPC_ref_tracking_pendulum}
\end{figure}

We test the performance of the neural adaptive MPC versus an implementation without adaptation, by modifying the physical parameters as follows: the first joint friction coefficient is changed from $c_1 = 0.1 Nms/rad$ to $c_1 = 0.01 Nms/rad$, while the mass of the second beam from $m_2 = 0.2 Kg$ to $m_2 = 0.3 Kg$. The performance of the two control algorithms, without and with adaptation, are reported in Figures \ref{fig: non adaptive MPC} and \ref{fig:Adaptive MPC}.
Finally, a new test has been carried out to control the double pendulum in an unstable position which is not an equilibrium point. The system's state has been initialized at $[0.1; 0.; -0.1; 0]$, while the initial estimated state has been set to zero. The reference has been set to $[0.7; 0; -1.4; 0]$. Figures \ref{fig: MPC_ref_tracking_pendulum} and \ref{fig: NAMPC_ref_tracking_pendulum} illustrate the results of the test. Note that the state estimation error is biased when not adapting the model due to the wrong identified system's gain.

\vspace{-0.1cm}
\section{Conclusions}\label{Conc}
Neural networks are a useful tool to obtain reliable system models for data-driven control. The use of these methods for control of unstable systems is still a largely open issue, manly due to the difficulty to train in such regimes. To this end, Tustin Networks (TN) have been proposed, which exploit the position/velocity relationship in mechanical systems. An adaptive TN MPC scheme  has been used for the control of a double pendulum with changing parameters and unmeasured velocity.

\bibliographystyle{plain}
\bibliography{mybiblio, semina}

\end{document}